\documentclass{article}

\pdfoutput=1

\usepackage[preprint]{neurips_2020}
\usepackage{natbib}
\usepackage[utf8]{inputenc} 
\usepackage[T1]{fontenc}    
\usepackage{hyperref}       
\usepackage{url}            
\usepackage{booktabs}       
\usepackage{amsfonts}       
\usepackage{nicefrac}       
\usepackage{microtype}      
\usepackage{subfigure}
\usepackage{amsmath,amsfonts,amssymb, amsthm}
\usepackage{bm}
\usepackage{hyperref}
\usepackage{url}
\usepackage{graphicx}
\usepackage{wrapfig}
\usepackage{physics}
\usepackage{placeins}
\usepackage{xcolor}
\usepackage{float}
\usepackage{flafter}
\usepackage{mathtools}
\usepackage{multirow}
\usepackage{bbold}

\title{Physics-guided Neural Networks (PGNN): \\ An Application in Lake Temperature Modeling}

\author{%
Arka Daw\thanks{Equal Contribution}\\
Virginia Tech\\
\texttt{darka@vt.edu}
\And
Anuj Karpatne\footnotemark[1]\\
Virginia Tech \\
\texttt{karpatne@vt.edu} \\
\AND
William Watkins \\
U.S. Geological Survey\\
\texttt{wwatkins@usgs.gov} \\
\And
Jordan Read\\
U.S. Geological Survey\\
\texttt{jread@usgs.gov} \\
\And
Vipin Kumar\\
University of Minnesota\\
\texttt{kumar001@umn.edu} \\
}

\renewcommand{\th}{%
    \ifmmode
        ^\mathrm{th}%
    \else%
        \textsuperscript{th}\xspace%
    \fi%
}

\begin{document}
\maketitle
\vspace{-4ex}
\begin{abstract}
This paper introduces a framework for combining scientific knowledge of physics-based models with neural networks to advance scientific discovery. This framework, termed physics-guided neural networks (PGNN), leverages the output of physics-based model simulations along with observational features in a hybrid modeling setup to generate predictions using a neural network architecture. Further, this framework uses physics-based loss functions in the learning objective of neural networks to ensure that the model predictions not only show lower errors on the training set but are also scientifically consistent with the known physics on the unlabeled set. We illustrate the effectiveness of PGNN for the problem of lake temperature modeling, where physical relationships between the temperature, density, and depth of water are used to design a physics-based loss function. By using scientific knowledge to guide the construction and learning of neural networks, we are able to show that the proposed framework ensures better generalizability as well as scientific consistency of results. All the code and datasets used in this study have been made available on this link \url{https://github.com/arkadaw9/PGNN}.

\end{abstract}

\section{Introduction}
\label{sec:intro}





Data science has become an indispensable tool for knowledge discovery in the era of big data, as the volume of data continues to explode in practically every research domain. Recent advances in data science such as deep learning have been immensely successful in transforming the state-of-the-art in a number of commercial and industrial applications such as natural language translation and image classification, using billions or even trillions of data samples. In light of these advancements, there is a growing anticipation in the scientific community to unlock the power of data science methods for accelerating scientific discovery \cite{Appenzeller16,graham2008big, jonathan2011special, sejnowski2014putting}.

However, a major limitation in using ``black-box'' data science models, that are agnostic to the underlying scientific principles driving real-world phenomena, is their sole dependence on the available labeled data, which is often limited in a number of scientific problems. In particular, a black-box data science model for a supervised learning problem can only be as good as the representative quality of the labeled data trained on. When the size of both the training and test sets are small, it is easy to learn spurious relationships that look deceptively good on both training and test sets (even after using standard methods for model evaluation such as cross-validation), but do not generalize well outside the available labeled data. A more serious concern with black-box applications of data science models is the lack of consistency of its predictions with respect to the known laws of physics (demonstrated in section \ref{sec:results}). Hence, even if a black-box model achieves somewhat more accurate performance but lacks the ability to adhere to mechanistic understandings of the underlying physical processes, it cannot be used as a basis for subsequent scientific developments. 


\begin{figure}
\centering
\includegraphics[width=0.6\textwidth]{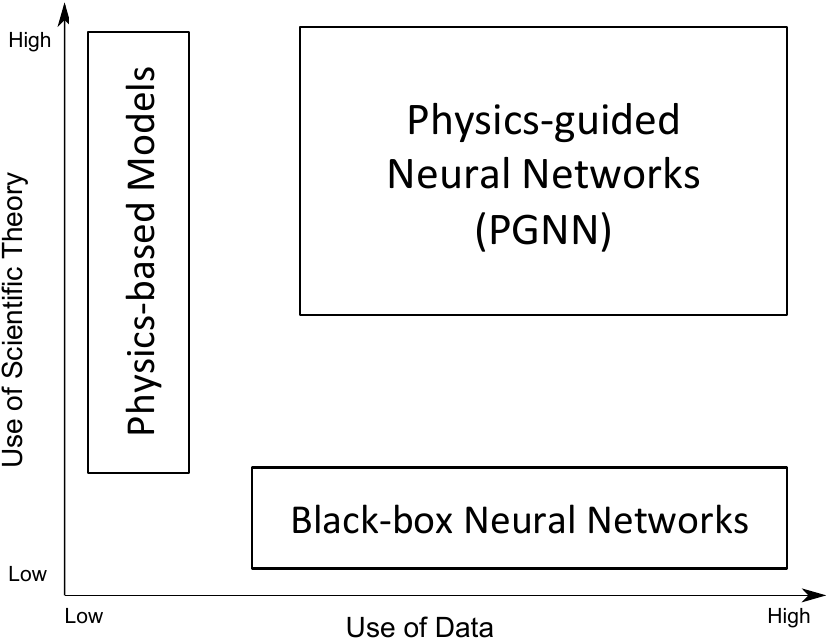}
\caption{A schematic representation of physics-guided neural networks in the context of other knowledge discovery approaches that either use physics or data. The $X$-axis measures the use of data while the $Y$-axis measures the use of scientific knowledge.}
\label{fig:tgds}
\end{figure}

On the other end of the spectrum, physics-based models,
which are founded on core scientific principles, strive to advance our understanding of the physical world by learning explainable relationships between input and output variables. These models have been the cornerstone of knowledge discovery in a wide range of scientific and engineering disciplines. There are two basic forms in which physical knowledge is generally available: (a) as physics-based rules or equations that dictate relationships between physical variables, and (b) in the form of numerical models of complex physical systems, e.g., simulations of dynamical systems that are heavily used in computational chemistry, fluid dynamics, climate science, and particle physics. While these models have significantly advanced our understanding of the physical universe, they are limited in their ability to extract knowledge directly from data and are mostly reliant only on the available physics. For example, many physics-based models use parameterized forms of approximations for representing complex physical processes that are either not fully understood or cannot be solved using computationally tractable methods.
Calibrating the parameters in physics-based models is a challenging task because of the combinatorial nature of the search space. In particular, this can result in the learning of over-complex models that lead to incorrect insights even if they appear interpretable at a first glance. For example, these and other challenges in modeling
hydrological processes using state-of-the-art physics-based models were the subject of  a series of debate papers in Water Resources Research (WRR) \cite{gupta2014debates, lall2014debates, mcdonnell2014debates}. One perspective \cite{gupta2014debates} argues that many physics-based models are excessively constrained by their \emph{a priori} parameterizations.
The dichotomy between physics-based models and black-box neural network models is schematically depicted in Figure \ref{fig:tgds}, where they both occupy the two extreme ends of knowledge discovery, either relying only on the data (black-box neural networks) or only on scientific knowledge (physics-based models).

In this paper, we introduce a framework of knowledge discovery in scientific problems that combines the power of neural networks with physics-based models, termed physics-guided neural networks (PGNN). 
There are two primary contributions of this work. First, we present an approach to create hybrid combinations of physics-based models and neural network architectures to make full use of both physics and data. Second, we present a novel framework for training neural network architectures using the knowledge contained in physics-based equations, to ensure the learning of physically consistent solutions. To demonstrate the framework of PGNN, we consider the illustrative problem of modeling the temperature of water in a lake at varying depths and times, using input drivers as well as physics-based model simulations. For this problem, we exploit a key physical relationship between the temperature, density, and depth of water in the form of physics-based loss function. 

The remainder of this paper is organized as follows. Section \ref{sec:pgnn} presents the generic framework of physics-guided neural networks that can be applied in any domain with some availability of scientific knowledge. Section \ref{sec:problem} presents the specific PGNN formulation for the illustrative problem of lake temperature modeling. Section \ref{sec:results} describes the evaluation procedure and presents experimental results, Section \ref{sec:discussions} presents some discussion on the approach used for hybrid modeling, while Section \ref{sec:conclusions} provides concluding remarks.

\section{Physics-guided Neural Networks}
\label{sec:pgnn}

The generic framework of physics-guided neural networks (PGNN) involves two key steps: (a) creating hybrid combinations of physics-based models and neural networks, termed hybrid-physics-data (HPD) models, and (b) using scientific knowledge as physics-based loss functions in the learning objective of neural networks, as described in the following.

\subsection{Constructing Hybrid-Physics-Data Models}
\label{sec:HPD}

Consider a predictive learning problem where we are given a set of input drivers, $\mathbf{D}$, that are physically related to a target variable of interest, $Y$.
A standard approach is to train a data science model, e.g., a neural network, $f_{NN}: \mathbf{D} \rightarrow Y$, over a set of training instances, which can then be used to produce estimates of the target variable, $\hat{Y}$. Alternatively, we can also use a physics-based numerical model, $f_{PHY}: \mathbf{D} \rightarrow Y$, to simulate the value of the target variable, $Y_{PHY}$, given its physical relationships with the input drivers. Analogous to the process of training, physics-based models often require ``calibrating" their model parameters using observational data---a process that is both time-consuming and label-expensive. Furthermore, $Y_{PHY}$ may provide an incomplete representation of the target variable due to simplified or missing physics in $f_{PHY}$, thus resulting in model discrepancies with respect to observations. Hence, the basic goal of HPD modeling is to combine $f_{PHY}$ and $f_{NN}$ so as to overcome their complementary deficiencies and leverage information in both physics and data.

\begin{figure}
\centering
\includegraphics[width=0.6\textwidth]{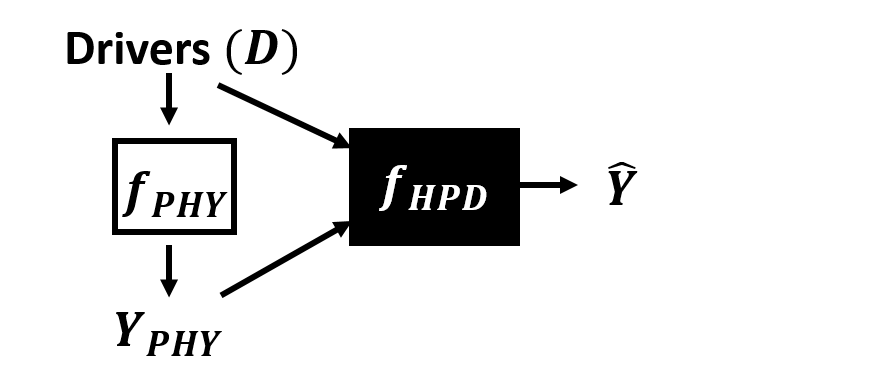}
\caption{A schematic illustration of a basic hybrid-physics-data (HPD) model, where the output $Y_{PHY}$ of a physics-based model $f_{PHY}$ is used as another feature in the data science model $f_{HPD}$ along with the drivers $D$ to produce the final outputs \^{Y}. In this schematic, white boxes represent physics-based models while black boxes represent ML models.}
\label{fig:hpd}
\end{figure}

One simple way for combining $f_{PHY}$ and $f_{NN}$ is to use the simulated outputs of the physics-based model, $Y_{PHY}$, as another input in the data science model (neural network) along with the drivers, $\mathbf{D}$. This results in the following basic HPD model: $$f_{HPD}: \mathbf{X} = [\mathbf{D}, ~Y_{PHY}] \rightarrow Y,$$
which is schematically illustrated in Figure \ref{fig:hpd}. In this setup, notice that if the physics-based model is accurate and $Y_{PHY}$ perfectly matches with observations of $Y$, then the HPD model can learn to predict $\hat{Y} = Y_{PHY}$. However, if there are systematic discrepancies (biases) in $Y_{PHY}$, then $f_{HPD}$ can learn to complement them by extracting complex features from the space of input drivers and thus reducing our knowledge gaps.


\subsection{Using Physics-based Loss Functions}
A standard approach for training the HPD model described in Figure \ref{fig:hpd} is to minimize the empirical loss of its model predictions, $\hat{Y}$, on the training set, while maintaining low model complexity as follows:
\begin{equation}
 \textup{arg} \min_{f} ~~ Loss(\hat{Y},Y) + \lambda ~ R(f), \label{ds_loss}
\end{equation}
where $R(.)$ measures the complexity of a model and $\lambda$ is a trade-off hyper-parameter. 
However, the effectiveness of any such training procedure is limited by the size of the labeled training set, which is often small in many scientific problems. In particular, there is no guarantee that model trained by minimizing Equation \ref{ds_loss} will produce results that are \emph{consistent} with our knowledge of physics. Hence, we introduce physics-based loss functions to guide the learning of data science models to physically consistent solutions as follows.

Let us denote the physical relationships between the target variable, $Y$, and other physical variables, $\mathbf{Z}$ using the following equations:
\begin{eqnarray}
\mathcal{G}(Y,\mathbf{Z}) &=& 0, \nonumber \\
\mathcal{H}(Y,\mathbf{Z}) &\leq& 0. \label{phy_eq}
\end{eqnarray}
Note that $\mathcal{G}$ and $\mathcal{H}$ are generic forms of physics-based equations that can either involve algebraic manipulations of $Y$ and $\mathbf{Z}$ (e.g., in the laws of kinematics), or their partial differentials (e.g., in the  Navier--Stokes equation for studying fluid dynamics or in the Schr\"{o}dinger equation for studying computational chemistry).  These physics-based equations must meet the same criteria as other loss function terms (i.e. continuous and differentiable). One way to measure if these physics-based equations are being violated in the model predictions, $\hat{Y}$, is to evaluate the following physics-based loss function:

\begin{equation}
    Loss.{PHY}(\hat{Y}) = ||\mathcal{G}(\hat{Y},\mathbf{Z})||^2 + \text{ReLU}~(\mathcal{H}(\hat{Y},\mathbf{Z})),
\end{equation}
where ReLU$(.)$ denotes the rectified linear unit function. Since $Loss.{PHY}$ does not require actual observations of the target variable, $Y$, it can be evaluated even on unlabeled data instances, in contrast to traditional loss functions. 
The complete learning objective of PGNN involving $Loss.{PHY}$ can then be stated as:

\begin{equation}
    \textup{arg} \min_{f} ~~ \underbrace{Loss(\hat{Y},Y)}_\text{Empirical Error} ~~~+ \underbrace{\lambda ~ R(f)}_\text{Structural Error} +~~~ \underbrace{\lambda_{PHY} ~ Loss.{PHY}(\hat{Y})}_\text{Physical Inconsistency}, \label{pgnn_loss}
\end{equation}

where $\lambda_{PHY}$ is the hyper-parameter that decides the relative importance of minimizing physical inconsistency compared to the empirical loss and the model complexity. Since the known laws of physics are assumed to hold equally well for any unseen data instance, ensuring physical consistency of model outputs as a learning objective in PGNN can help in achieving better generalization performance even when the training data is small and not fully representative. Additionally, the output of a PGNN model can also be interpreted by a domain expert and ingested in scientific workflows, thus leading to scientific advancements. 

There are several optimization algorithms that can be used for minimizing Equation \ref{pgnn_loss}, e.g., the stochastic gradient descent (SGD) algorithm and its variants that have found great success in training deep neural networks. In particular, the gradients of $Loss.{PHY}$ w.r.t model parameters can be easily computed using the automatic differentiation procedures available in standard deep learning packages. This makes neural networks a particularly suited choice for incorporating physics-based loss functions in the learning objective of data science models.

\section{PGNN for Lake Temperature Modeling}
\label{sec:problem}

In this section, we describe our PGNN formulation for the illustrative problem of modeling the temperature of water in lakes. In the following, we first provide some background information motivating the problem of lake temperature modeling, and then describe our PGNN approach.

\subsection{Background: Lake Temperature Modeling}

The temperature of water in a lake is known to be an ecological ``master factor'' \cite{magnuson1979temperature} that controls the growth, survival, and reproduction of fish (e.g., \cite{roberts2013fragmentation}). 
Warming water temperatures can increase the occurrence of aquatic invasive species \cite{rahel2008assessing,roberts2017nonnative}, which may displace fish and native aquatic organisms, and result in more harmful algal blooms (HABs) \cite{harris2017predicting,paerl2008blooms}.
Understanding temperature change and the resulting biotic ``winners and losers'' is timely science that can also be directly applied to inform priority action for natural resources. Accurate water temperatures (observed or modeled) are critical to understanding contemporary change, and for predicting future thermal habitat of economically valuable fish. 

Since observational data of water temperature at broad spatial scales is incomplete (or non-existent in some regions) high-quality temperature modeling is necessary. 
Of particular interest is the problem of modeling the temperature of water at a given depth\footnote{Depth is measured in the direction from the surface of the water to the lake bottom.}, $d$, and on a certain time, $t$. This problem is referred to as 1D-modeling of temperature (depth being the single dimension). A number of physics-based models have been developed for studying lake temperature, e.g., the state-of-the-art general lake model (GLM) \cite{hipsey2014glm}.
\begin{figure}
\centering
\includegraphics[width=0.8\linewidth]{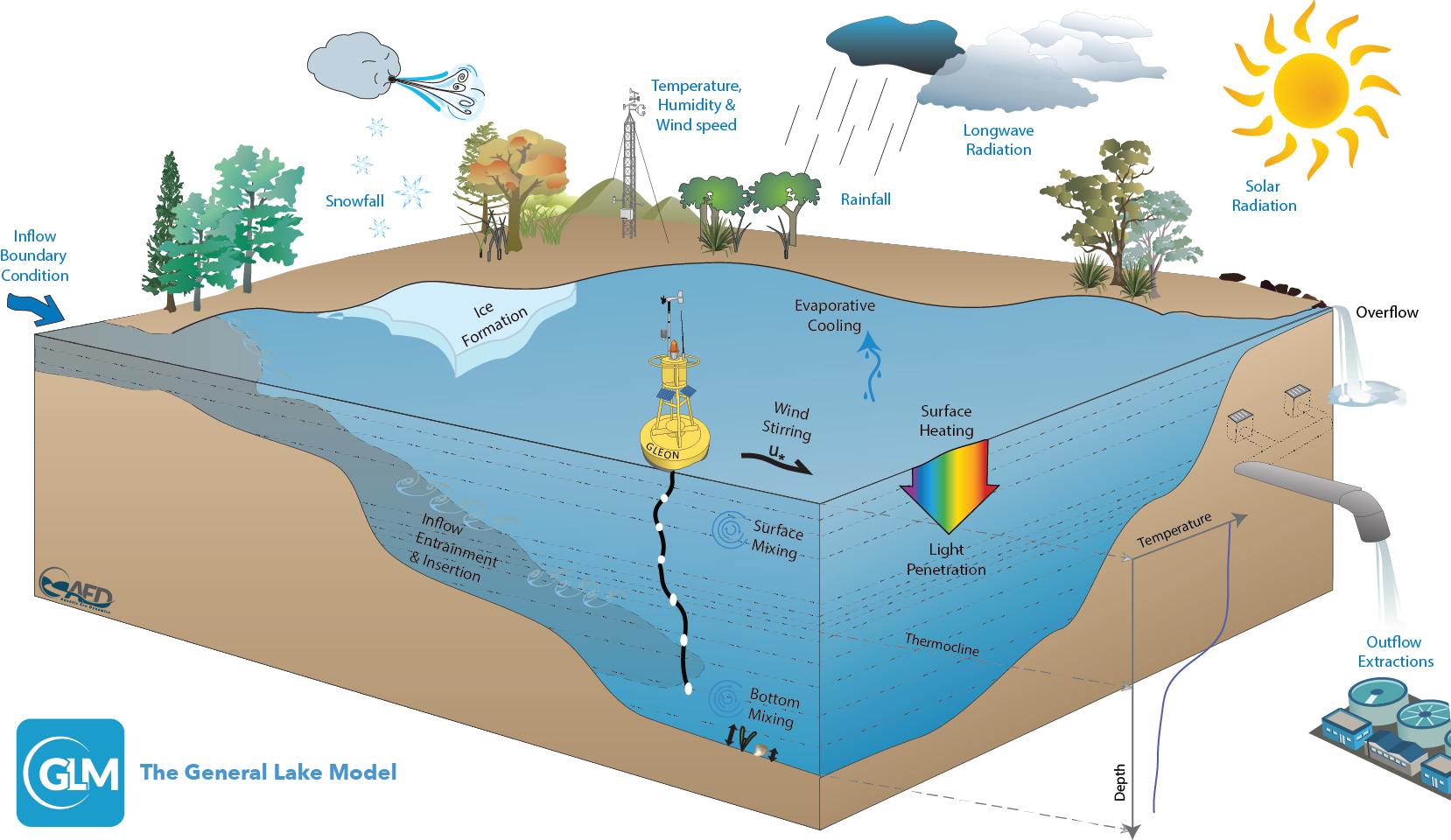}
\caption{A pictorial description of the physical processes governing the dynamics of temperature in a lake. Figure courtesy: \cite{hipsey2014glm}. (Note: Figures in this paper are best viewed in color.)}
\label{fig:glm}
\end{figure}
This model  captures a variety of physical processes governing the dynamics of temperature in a lake, e.g., the heating of the water surface due to incoming shortwave radiation from the sun, the attenuation of radiation beneath the surface and the mixing of layers with varying energies at different depths, and the dissipation of heat from the surface of the lake via evaporation or longwave radiation, shown pictorially in Figure \ref{fig:glm}. We use GLM as our preferred choice of physics-based model for lake temperature modeling.

The GLM has a number of parameters (e.g., parameters related to vertical mixing, wind energy inputs, and water clarity) that needs to be custom-calibrated for each lake if some training data is available. 
The basic idea behind these calibration steps is to run the model for each possible combination of parameter values and select the one that has maximum agreement with the observations. Because this step of custom-calibrating is both labor- and computation-intensive, there is a trade-off between increasing the accuracy of the model and expanding the feasability of study to a large number of lakes. 

\subsection{Proposed PGNN Formulation}

We consider the physical variables governing the dynamics of lake temperature at every depth and time-step as the set of input drivers, $\mathbf{D}$. This includes meteorological recordings at the surface of water such as the amount of solar radiation at different wavelengths, wind speed, and air temperature, as well as the value of depth and the day of the year. To construct an HPD model of the type shown in Figure \ref{fig:hpd}, we use simulations of lake temperature from the GLM, $Y_{PHY}$, along with the input drivers $\mathbf{D}$ at every depth and time-step to obtain the augmented set of features, $$\mathbf{X} = [\mathbf{D},~ Y_{PHY}].$$

We adopt a basic multi-layer perceptron architecture to regress the temperature, $Y$, on any given depth and time, using $\mathbf{X}$. For a fully-connected network with $L$ hidden layers, this amounts to the following modeling equations relating the input features, $\mathbf{x}$, to its target prediction, $\hat{y}$:
\begin{eqnarray}
    \mathbf{z}_1 &=& \mathbf{W}_1^T \mathbf{x} + \mathbf{b}_1 \\
    \mathbf{z}_i &=& \mathbf{W}_i^T \mathbf{a}_{i-1} + \mathbf{b}_i \quad \forall ~ i = 2 ~\text{to}~ L \\
    \mathbf{a}_i &=& f(\mathbf{z}_i) \quad \forall ~ i = 1 ~\text{to}~ L \\
    \hat{y} &=& \mathbf{w}_{L+1}^T \mathbf{a}_{L} + {b}_{L+1}
\end{eqnarray}
where $(\mathbf{W},\mathbf{b}) = \{(\mathbf{W}_i,\mathbf{b}_i)\}_1^{L+1}$ represents the set of weight and bias parameters across all hidden and output layers, and $f$ is the activation function used at the hidden layers.
We use the mean squared error as our choice of loss function and $L_1$ and $L_2$ norms of network weights, $\mathbf{W}$ as regularization terms in Equation \ref{ds_loss} as follows:
\begin{eqnarray}
    Loss(\hat{Y},Y) &=& \frac{1}{n} \sum_{i=1}^n (y_i - \hat{y_i})^2, \label{mse_loss} \\
    \lambda ~ R(\mathbf{W}) &=&  \lambda_1 ||\mathbf{W}||_1 + + \lambda_{2} ||\mathbf{W}||_2, \label{reg_term}
\end{eqnarray}
where $\{\mathbf{x},y\}_{1}^{n}$ is the set of training instances.

To incorporate the knowledge of physics as a loss function in the training of neural networks, we employ a key physical relationship between the temperature, density, and depth of water as our physics-based equation (Equation \ref{phy_eq}). In the following, we introduce the two key components of this physical relationship and describe our approach for using it to ensure the learning of physically consistent results.

\begin{figure}[tb]
\centering
\subfigure[Temperature--Density Relationship]{\label{fig:temp-density} \includegraphics[width=0.45\textwidth]{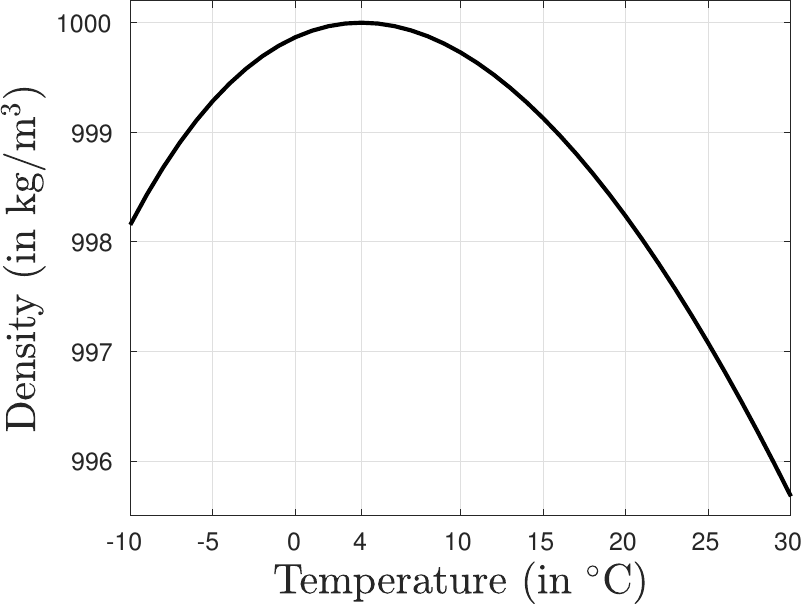}}
\quad
\subfigure[Density--Depth Relationship]{\label{fig:density-depth} \includegraphics[width=0.3\textwidth]{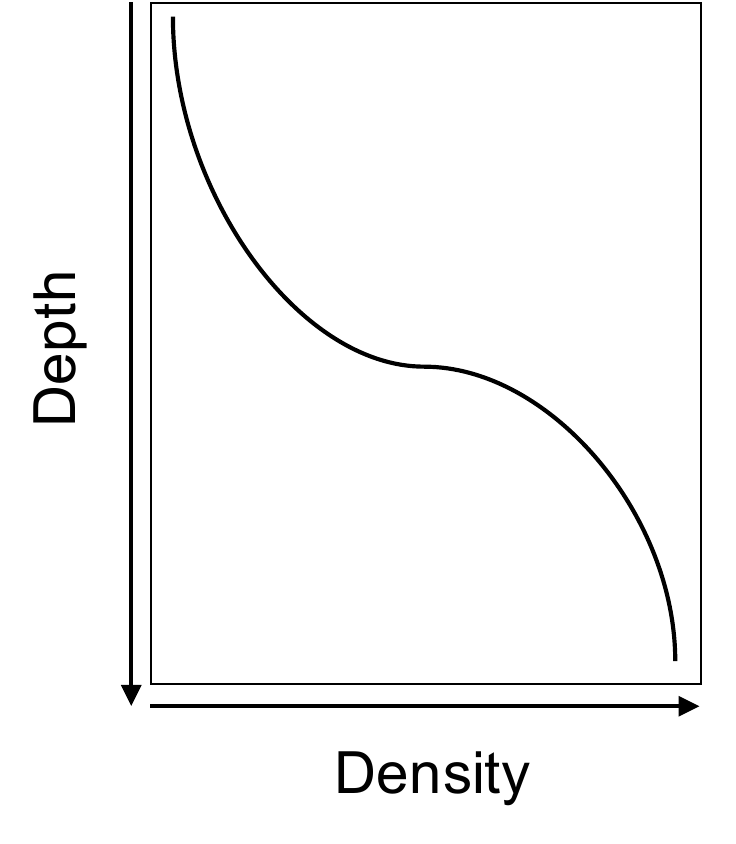}}
\caption{Plots of physical relationships between temperature, density, and depth of water that serve as the basis for introducing physical consistency in PGNN.}
\label{fig:phy_rel}
\end{figure}

\subsubsection{Temperature--Density Relationship:}
The temperature, $Y$, and density, $\rho$, of water are non-linearly related to each other according to the following known physical equation \cite{martin1998hydrodynamics}:

\begin{equation}
\rho = 1000 \times \Big( 1 - \frac{(Y + 288.9414) \times (Y - 3.9863)^2}{508929.2 \times (Y + 68.12963) } \Big)
\label{eq:temp-density}
\end{equation}

Figure \ref{fig:temp-density} shows a plot of this relationship between temperature and density, where we can see that water is maximally dense at 4$^\circ$Celsius (due to the hydrogen bonding between water molecules)\footnote{This simple fact is responsible for the sustenance of all forms of aquatic life on our planet, as water at 4$^\circ$C moves down to the bottom and stops the freezing of lakes and oceans.}. 
Given the temperature predictions of a model, $\hat{Y}[d,t]$, at depth, $d$, and time-step, $t$, we can use Equation \ref{eq:temp-density} to compute the corresponding density prediction, $\hat{\rho}[{d,t}]$.

\subsubsection{Density--Depth Relationship:}
The density of water monotonically increases with depth as shown in the example plot of Figure \ref{fig:density-depth}, since denser water is heavier and goes down to the bottom of the lake. Formally, the density of water at two different depths, $d_1$ and $d_2$, on the same time-step, $t$, are related to each other in the following manner:
\begin{equation}
    \rho[d_1,t] - \rho[d_2,t] \leq 0 \quad \text{if}~ d_1 < d_2.
    \label{eq:density-depth}
\end{equation}

To ensure that this physics-based equation is upheld in the temperature predictions of a physics-based model, $\hat{Y}$, we can construct a physics-based loss function as follows. Let us consider an unlabeled data set of input features on a regular grid of $n_d$ depth values and $n_t$ time-steps. On any pair of consecutive depth values, $d_i$ and $d_{i+1}$ ($d_{i} < d_{i+1}$), we can compute the difference in the density estimates of a model on time-step $t$ as
\begin{equation}
    \Delta[i,t] = \hat{\rho}[d_i,t] - \hat{\rho}[d_{i+1},t]
\end{equation}
A positive value of $\Delta[i,t]$ can be viewed as a violation of the physics-based equation \ref{eq:density-depth} on depth $d_i$ and time $t$. This can be evaluated as a non-zero occurrence of $\text{ReLU}(\Delta[d_i,t])$. Hence, we can consider the mean of all physical violations across every consecutive depth-pair and time-step as our physics-based loss function:
\begin{equation}
    {PHY}.Loss(\hat{Y}) = \frac{1}{n_t ({n_d} - 1)} \sum_{t=1}^{n_t} \sum_{i=1}^{{n_d} -1} \text{ReLU}(\Delta[{i,t}]).
    \label{eq:physics_loss}
\end{equation}
Using this physics-based loss (Equation \ref{eq:physics_loss}) along with the empirical loss (Equation \ref{mse_loss}) and regularization terms (Equation \ref{reg_term}) in the learning objective (Equation \ref{pgnn_loss}), we obtain our complete PGNN formulation.
Note that in our particular problem of lake temperature modeling, even though the neural network is being trained to improve its accuracy on the task of predicting water temperatures, the use of physics-based loss function ensures that the temperature predictions also translate to consistent relationships between other physical variables, namely density and depth, thus resulting in a wholesome solution to the physical problem.



\section{Evaluation}
\label{sec:results}

In this section, we first describe the data collected over two lakes for evaluation along with the experimental design, choice of baselines, evaluation metrics, and experimental results.

\subsection{Data}
We consider two example lakes to demonstrate the effectiveness of our PGNN framework for lake temperature modeling, Mille Lacs Lake in Minnesota, USA, and Lake Mendota in Wisconsin, USA. Both these lakes are reasonably large (536 km$^2$ and 40 km$^2$ in area, respectively), have extensive observation records relative to other similar lakes, and show sufficient dynamics in the temperature profiles across depth over time to make them interesting test cases for analyses. 
Observations of lake temperature were collated from a variety of sources including Minnesota Department of Natural Resources and a web resource that collates data from federal and state agencies, academic monitoring campaigns, and citizen data \cite{read2017water}. These temperature observations vary in their distribution across depths and time, with some years and seasons being heavily sampled, while other time periods having little to no observations.

The overall data for Mille Lacs Lake consisted of 7,072 temperature observations from 17 June 1981 to 01 Jan 2016, and the overall data for Lake Mendota consisted of 13,543 temperature observations from 30 April 1980 to 02 Nov 2015. For each observation, we used a set of 11 meteorological drivers as input variables, listed in Table \ref{tab:inputs}. While many of these drivers were directly measured, we also used some domain-recommended ways of constructing derived features such as Growing Degree Days \cite{prentice1992special}.
We used the General Lake Model (GLM) \cite{hipsey2014glm} as the physics-based approach for modeling lake temperature in our experimental studies. The GLM uses the drivers listed in Table \ref{tab:inputs} as input parameters and balances the energy and water budget of lakes or reservoirs on a daily or sub-daily timestep. 
It performs a 1D modeling (along depth) of a variety of lake variables (including water temperature) using a vertical Lagrangian layer scheme. 

Apart from the labeled set of data instances where we have observations of temperature, we also considered a large set of unlabeled instances (where we do not have temperature observations) on a regular grid of depth values at discrete steps of 0.5m, and on a daily time-scale from 02 April 1980 to 01 Jan 2016 (amounting to 13,058 dates). We ran the GLM model on the unlabeled instances to produce $Y_{PHY}$ along with the input drivers $\mathbf{D}$ at every unlabeled instance. Ignoring instances with missing values, this amounted to a total of 299,796 unlabeled instances in Mille Lacs Lake and 662,781 unlabeled instances in Lake Mendota.

\begin{table}[h]
\centering
\begin{tabular}{l|l|}
\cline{2-2}
                         & \multicolumn{1}{c|}{Input Drivers} \\ \hline
\multicolumn{1}{|l|}{1}  & Day of Year (1 -- 366)              \\ \hline
\multicolumn{1}{|l|}{2}  & Depth (in m)                       \\ \hline
\multicolumn{1}{|l|}{3}  & Short-wave Radiation (in W/m$^2$)              \\ \hline
\multicolumn{1}{|l|}{4}  & Long-wave Radiation   (in W/m$^2$)              \\ \hline
\multicolumn{1}{|l|}{5}  & Air Temperature (in $^\circ C$)                    \\ \hline
\multicolumn{1}{|l|}{6}  & Relative Humidity (0 -- 100 \%)               \\ \hline
\multicolumn{1}{|l|}{7}  & Wind Speed (in m/s)                        \\ \hline
\multicolumn{1}{|l|}{8} & Rain (in cm)         \\ \hline
\multicolumn{1}{|l|}{9}  & Growing Degree Days  \cite{prentice1992special}              \\ \hline
\multicolumn{1}{|l|}{10}  & Is Freezing (True or False)        \\ \hline
\multicolumn{1}{|l|}{11} & Is Snowing (True or False)         \\ \hline
\end{tabular}
\caption{Input drivers for lake temperature modeling.}
\label{tab:inputs}
\end{table}

\subsection{Experimental Design}
We considered contiguous windows of time to partition the labeled data set into training and test splits, to ensure that the test set is indeed independent of the training set and the two data sets are not temporally auto-correlated. In particular, we chose the center portion of the overall time duration for testing, while the remainder time periods on both ends were used for training. For example, to construct a training set of $n$ instances, we chose the median date in the overall data and kept on adding dates on both sides of this date for testing,  till the number of observations in the remainder time periods became less than or equal to $n$. Using this protocol, we constructed training sets of size $n = 3000$ for both Mille Lacs Lake and Lake Mendota, which were used for calibrating the physics-based model, PHY, on both lakes. We used the entire set of unlabeled instances for evaluating the physics-based loss function on every lake.


All neural network models used in this paper were implemented using the Keras package \cite{chollet2015} using Tensorflow backend. We used the AdaDelta algorithm \cite{zeiler2012adadelta} for performing stochastic gradient descent on the model parameters of the neural network. We used a batch size of 1000 with maximum number of epochs equal to 10,000. To avoid over-fitting, we employed an early stopping procedure using  10\% of the training data for validation, where the value of patience was kept equal to 500. We also performed gradient clipping (for gradients with $L_2$ norm greater than 1) to avoid the problem of exploding gradients common in regression problems (since the value of $Y$ is unbounded). We standardized each dimension of the input attributes to have 0 mean and 1 standard deviation, and applied the same transformation on the test set. The fully-connected neural network architecture comprised of 3 hidden layers, each with 12 hidden nodes. The value of hyper-parameters $\lambda_1$ and $\lambda_2$ (corresponding to the $L_1$ and $L_2$ norms of network weights, respectively) were kept equal to 1 in all experiments conducted in the paper, to demonstrate that no special tuning of hyper-parameters was performed for any specific problem. The value of the hyper-parameter $\lambda_{PHY}$ corresponding to the physics-based loss function was kept equal to $std(Y^2)/std(\rho)$, to factor in the differences in the scales of the physics-based loss function and the mean squared error loss function. We used uniformly random initialization of neural network weights from 0 to 1. Hence, in all our experiments, we report the mean and standard deviation of evaluation metrics of every neural network method over 50 runs, each run involving a different random initialization.



\subsection{Baseline Methods and Evaluation Metrics}
We compared the results of PGNN with the following baseline methods:
\begin{itemize}
    \item \textbf{PHY}: The GLM models calibrated on the training sets of size $n = 3000$ for both lakes were used as the physics-based models, PHY. 
    \item \textbf{Black-box Models}: In order to demonstrate the value in incorporating the knoweldge of physics with data science models, we consider three standard non-linear regression models: support vector machine (\textbf{SVM}) with radial basis function (RBF) kernel, least squares boosted regression trees (\textbf{LSBoost}), and the neural network (\textbf{NN}) model. All of these models were trained to predict temperature using the same set of input drivers as PGNN, but without using any knowledge of physics (either in the form of model simulations or as physics-based loss functions).
    \item \textbf{PGNN0}: In order to understand the contribution of the physics-based loss function in PGNN, we consider an intermediate product of our framework, PGNN0, as another baseline, which uses the hybrid-physics-data modeling setup described in Figure \ref{fig:hpd}, but does not use the physics-based loss function in its learning objective (Equation \ref{ds_loss}). Hence, PGNN0 differs from black-box models in its use of physics-based model simulations as input attributes, and differs from PGNN in its use of a purely data-driven learning objective.
\end{itemize}

We considered the following evaluation metrics for comparing the performance of different algorithms:
\begin{itemize}
    \item \textbf{RMSE}: We use the root mean squared error (RMSE) of a model on the test set as an estimate of its generalization performance. The units of this metric are in $^\circ$C.
    \item \textbf{Physical Inconsistency}: Apart from ensuring generalizability, a key contribution of PGNN is to ensure the learning of physically consistent model predictions. Hence, apart from computing the RMSE of the model on the test set, we also compute the fraction of time-steps where the model makes physically inconsistent predictions (i.e., the density-depth relationship stated in Equation \ref{eq:density-depth} is violated). We report this fraction as the physical inconsistency measure in Figures \ref{fig:scatter}, \ref{fig:vary_instab}, and \ref{fig:phyinconvslambda}. Note that this measure does not require actual observations, and hence, we compute this measure over the plentifully large unlabeled data set.
\end{itemize}

\subsection{Results}

\begin{figure}[tb]
\centering
\subfigure[Results on Mille Lacs Lake]{\label{fig:ml_scatter} \includegraphics[width=0.45\textwidth]{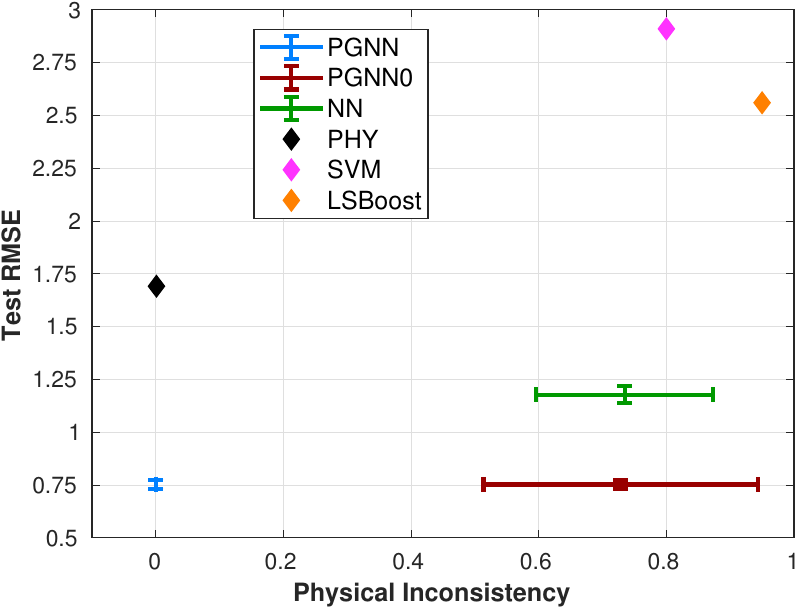}}
\subfigure[Results on Lake Mendota]{\label{fig:men_scatter} \includegraphics[width=0.45\textwidth]{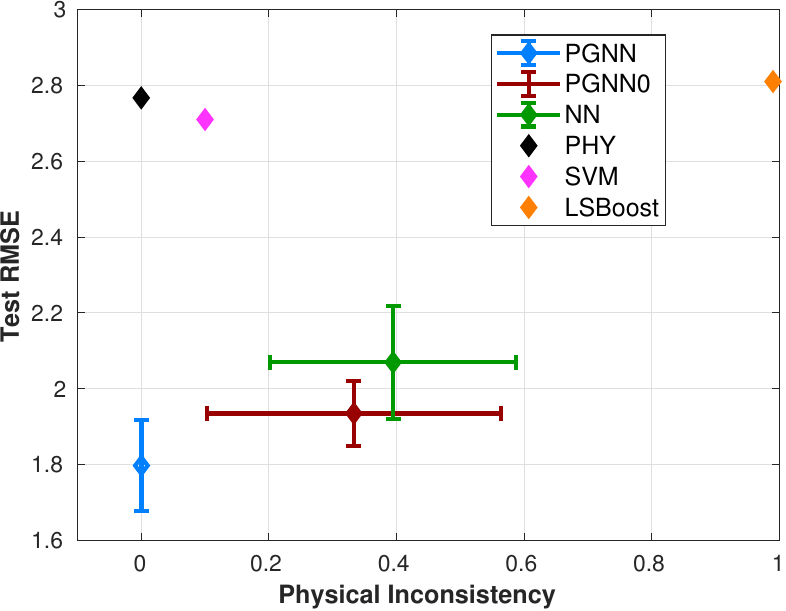}}
\caption{Scatter plots showing test RMSE values ($Y$-axis) and physical inconsistency ($X$-axis) of comparative methods. Points and error bars respectively represent the mean and +/- one standard deviation from the mean of results from all 50 random weight initializations.}
\label{fig:scatter}
\end{figure}

Figure \ref{fig:scatter} provides a summary of the performance of different methods for modeling lake temperature on the two example lakes, Mille Lacs Lake and Lake Mendota. The $X$-axis in these plots represents the physical inconsistency of a model, while the $Y$-axis represents the RMSE of the model predictions w.r.t. observations on the test set. We also show the standard deviation around the evaluation metrics of neural network-based methods (i.e., PGNN, PGNN0, and NN), since we used random initialization of network weights for every one of the 50 runs. 

For Mille Lacs Lake, we can see from Figure \ref{fig:ml_scatter} that the test RMSE of the physics-based model, PHY, is 1.69. If we use black-box data science models such as SVM and LSBoost, that try to learn non-linear relationships between drivers and temperature directly without using physics, we would end up with a test RMSE that is even higher than that of PHY. Further, they also show high physical inconsistency in their model predictions (greater than 0.8). If we instead use a black-box NN model that learns non-linear compositions of features from the space of input drivers, we can achieve a test RMSE of 1.18 that is significantly lower than that of PHY. This provides evidence of the information contained in the driver data, which if used effectively, can help in closing the knowledge gaps of PHY.
However, this improvement in RMSE comes at the cost of a large value of physical inconsistency in the model predictions of NN (almost 73\% of the time-steps have inconsistent density-depth relationships in its predictions). This makes NN unfit for use in the process of scientific discovery, because although it is able to somewhat improve the predictions of the target variable (i.e. temperature), it is incurring large errors in capturing the physical relationships of temperature with other variables, leading to non-meaningful results. 

If we use the output of the physics-based model along with the drivers as inputs  in the PGNN0 model, we can achieve an even lower value of test RMSE than that of NN. This is because the output of PHY (although with a high RMSE) contains vital physical information about the dynamics of lake temperature, which when coupled with powerful data science frameworks such as neural networks, can result in major improvements in RMSE. However, the results of PGNN0 are still physically inconsistent for roughly 72\% of the time. In contrast, it is only by the use of physics-based loss functions in PGNN that we can not only achieve an RMSE of 0.73, but also substantially lower  value of physical inconsistency (close to 0). To appreciate the significance of a drop in RMSE of 0.96$^\circ$C, note that a lake-specific calibration approach that produced a median RMSE of 1.47$^\circ$C over 28 lakes is considered to be the state-of-the-art in the field \cite{fang2012simulations}. By being accurate as well as physically consistent, PGNN provides an opportunity to produce physically meaningful analyses of lake temperature dynamics that can be used in subsequent scientific studies.

A similar summary of results can also be obtained from Figure \ref{fig:men_scatter} for Lake Mendota. We can see that the test RMSE of the physics-based model in this lake is 2.77, which is considerably higher than that of Mille Lacs Lake. This shows the relatively complex nature of temperature dynamics in Lake Mendota compared to Mille Lacs Lake, which are more difficult for any model to approximate. Mille Lacs Lake is generally well-mixed (i.e. bottom temperature is similar to the surface temperature) while Lake Mendota is more stratified.  The average test RMSE scores of NN and PGNN0 for Lake Mendota are 2.07 and 1.93, respectively. On the other hand, PGNN is able to achieve an average RMSE of 1.79, while being physically consistent. This is a demonstration of the added value of using physical consistency in the learning objective of data science models for improving generalization performance.

\subsubsection{Effect of Varying Training Size}

\begin{figure}[tb]
\centering
\subfigure[Effect on Test RMSE]{\label{fig:vary_rmse} \includegraphics[width=0.52\textwidth]{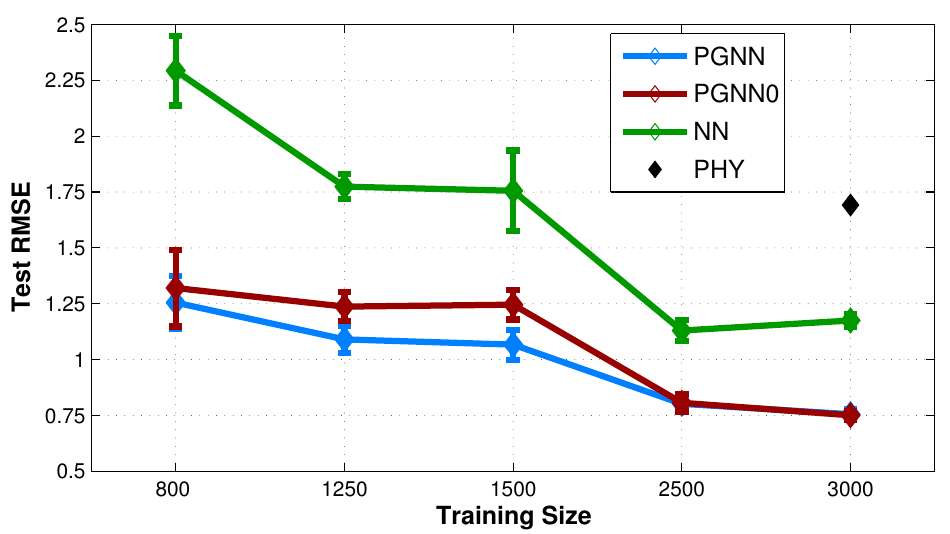}}
\subfigure[Effect on Physical Inconsistency]{\label{fig:vary_instab} \includegraphics[width=0.44\textwidth]{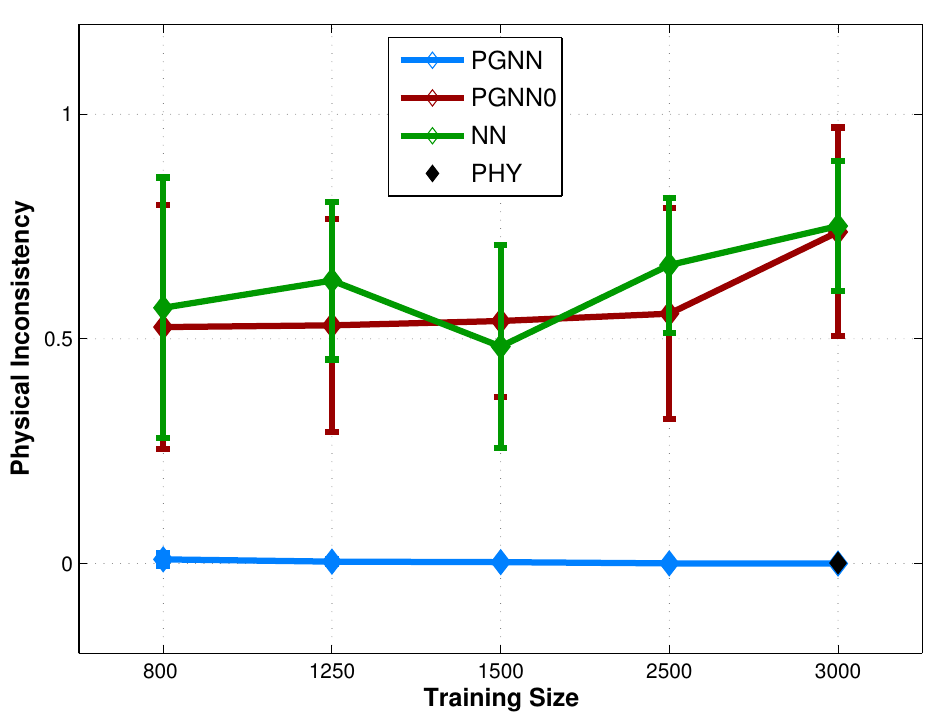}}
\caption{Effect of varying training size on the performance of different methods on Mille Lacs Lake.  Points and error bars respectively represent the mean and +/- one standard deviation from the mean of results from all 50 random weight initializations.}
\label{fig:vary_trsize}
\end{figure}

We next demonstrate the effect of varying the size of the training set on the performance of PGNN, in comparison with other baseline methods. Figure \ref{fig:vary_trsize} shows the variations in the test RMSE and physical inconsistency of different methods on Mille Lacs Lake, as we vary the training size from 3000 to 800. We can see from Figure \ref{fig:vary_rmse} that the test RMSE values of all data science methods increase as we reduce the training size. For example, the test RMSE of the black-box model, NN, can be seen to over-shoot the test RMSE of the physics-based model for training sizes smaller than 1500. On the other hand, both PGNN and PGNN0 show a more gradual increase in their test RMSE values on reducing training size. In fact, the PGNN can be seen to provide smaller RMSE values than all baseline methods, especially at training sizes of 1250 and 1500. This is because the use of physics-based loss function ensures that the learned PGNN model is consistent with our knowledge of physics and thus is not spurious. Such a model thus stands a better chance at capturing generalizable patterns and avoiding the phenomena of over-fitting, even after being trained with limited number of training samples. If we further reduce the training size to 800, the results of PGNN and PGNN0 become similar because there is not much information left in the data that can provide improvements in RMSE.

While the lower RMSE values of PGNN is promising, the biggest gains in using PGNN arise from its drastically lower values of physical inconsistency as compared to other data science methods, as shown in Figure \ref{fig:vary_instab}, even when the training sizes are small. Note that the results of PGNN are physically consistent across all time-steps, while PGNN0 and NN violate the density-depth relationship more than 50\% of time-steps on an average. We can also see that PHY has an almost zero value of physical inconsistency, since it is inherently designed to be physically consistent. 

\subsubsection{Sensitivity to hyperparameter $\mathbf{\lambda_{PHY}}$}

\begin{figure}[h]
\centering
\subfigure[Effect on Physical Inconsistency]{\label{fig:phyinconvslambda} \includegraphics[width=0.4\textwidth]{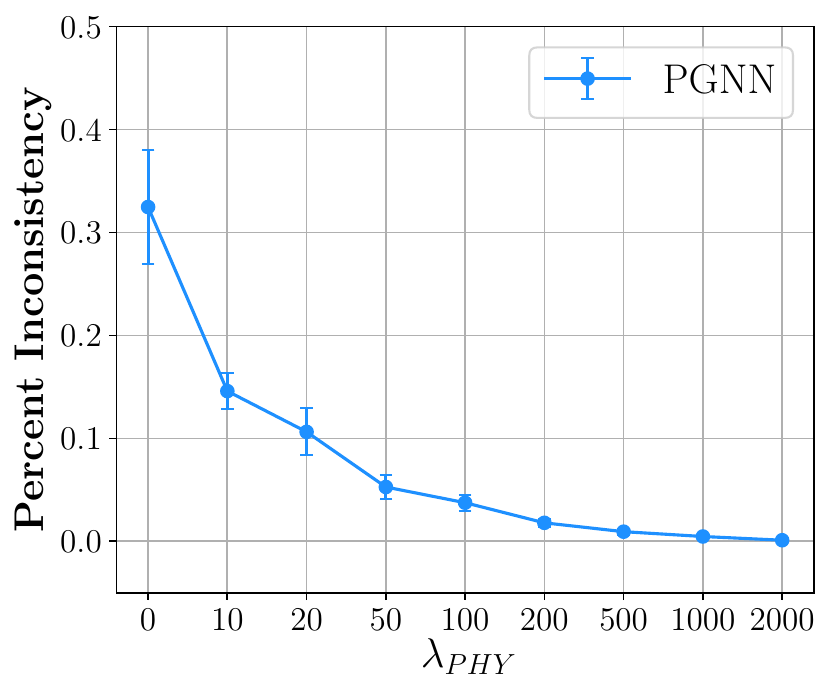}}
\subfigure[Effect on Test RMSE]{\label{fig:rmsevslambda} \includegraphics[width=0.4\textwidth]{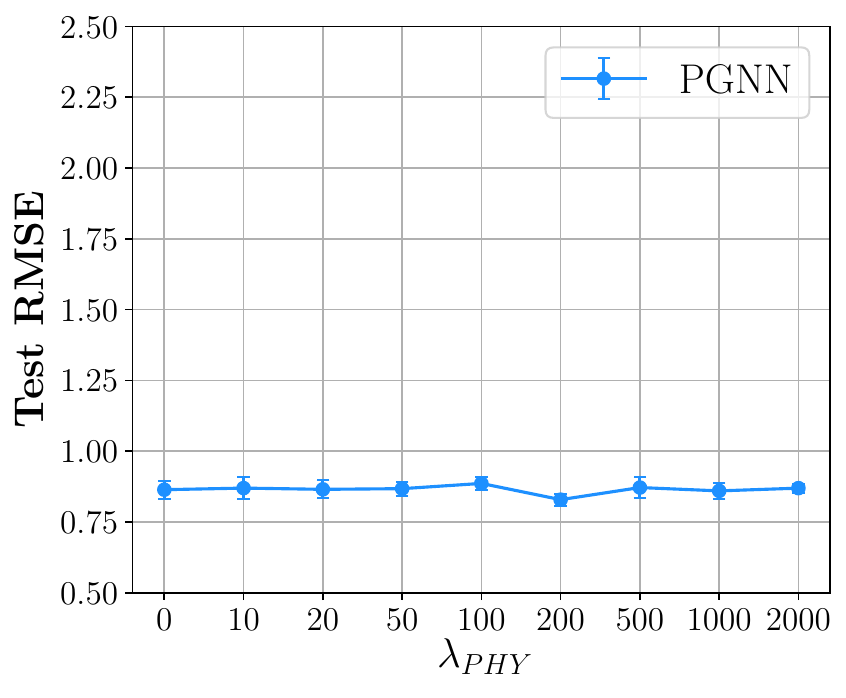}}
\caption{Sensitivity to hyperparameter $\lambda_{PHY}$ on Mille Lacs Lake. Points and error bars respectively represent the mean and +/- one standard deviations from the mean of results from all 50 random weight initializations.}
\label{fig:sensitivitylambda}
\end{figure}

To understand how the choice of the trade-off hyperparameter $\lambda_{PHY}$ affects the model results, we analyse the physical inconsistency and the Test RMSE while varying $\lambda_{PHY}$ (See Figure \ref{fig:sensitivitylambda}). With the increase in the value of $\lambda_{PHY}$, we impose a more stringent physics-constraint on the model which ultimately leads to the generation of more and more physically consistent predictions (Figure \ref{fig:phyinconvslambda}). Simultaneously, it can be observed that the change in $\lambda_{PHY}$ does not significantly affect the Test RMSE of the learned model which is also desirable (Figure \ref{fig:rmsevslambda}). Ideally, with the introduction of the physics-based loss during training, we would want the model to generate more physically consistent predictions while not degrading its predictive performance.

\subsubsection{Analysis of Results}
\begin{figure}[tb]
\centering
\subfigure[Mille Lacs Lake on 02-October-2012]{\label{fig:ml_profile} \includegraphics[width=0.4\textwidth]{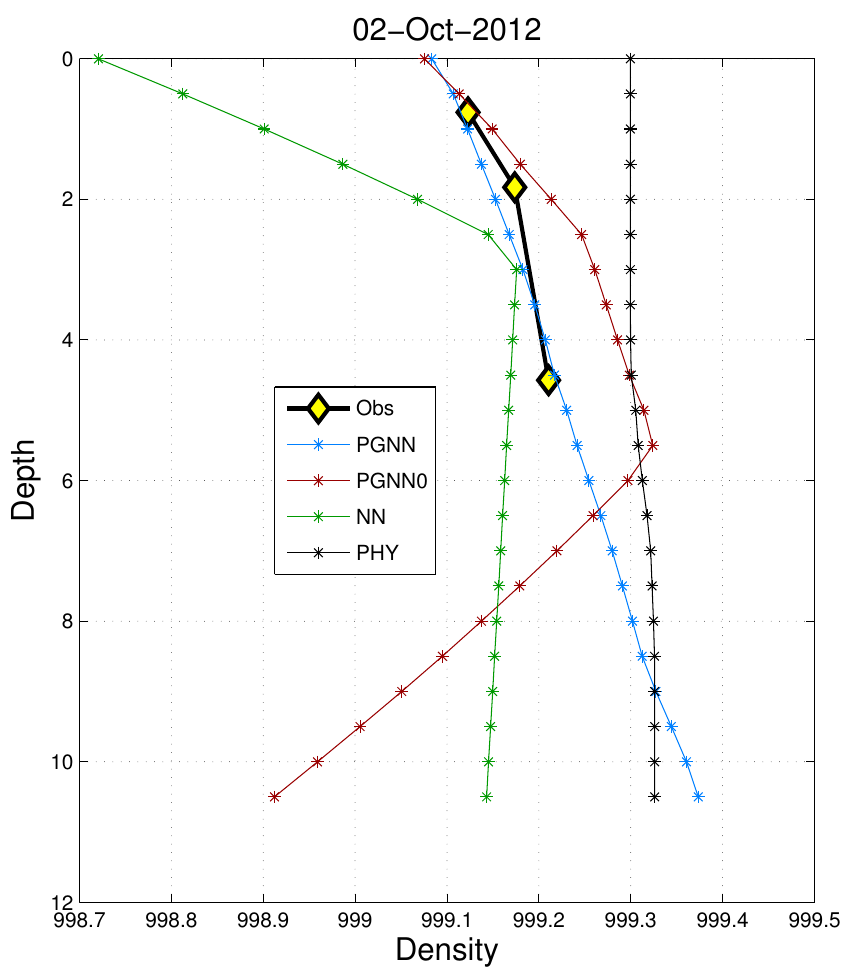}}
\subfigure[Lake Mendota on 27-May-2003]{\label{fig:men_profile} \includegraphics[width=0.4\textwidth]{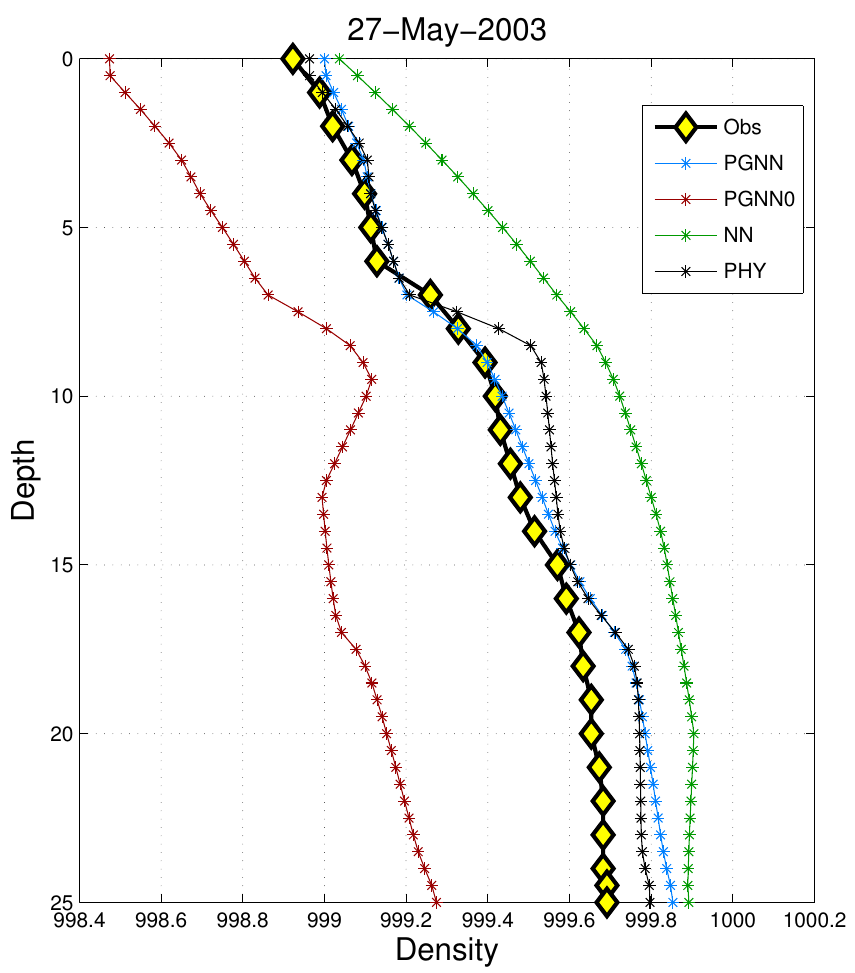}}
\caption{Density profiles of varying algorithms on different dates in Mille Lacs Lake (Figure \ref{fig:ml_profile}) and Lake Mendota (Figure \ref{fig:men_profile}).}
\label{fig:density_profiles}
\end{figure}

To provide a deeper insight into the results produced by competing methods, we analyze the predictions of lake temperature produced by a model as follows. As described previously, any estimate of temperature can be converted to its corresponding density estimate using the physical relationship between temperature and density represented in Equation \ref{eq:temp-density}. Hence, on any given time-step, we can produce a profile of density estimates at varying values of depth for every model, and match it with the density estimates of observed temperature on test instances. Visualizing such density profiles can help us understand the variations in model predictions across depth, in relationship to test observations. Some examples of density profiles on different dates in Mille Lacs Lake and Lake Mendota are provided in Figure \ref{fig:density_profiles}, where the $X$-axis represents estimated density, and the $Y$-axis represents depth.

In the density profiles of different algorithms on Mille Lacs Lake in Figure \ref{fig:ml_profile}, we can see that the density estimates of PHY are removed from the actual observations by a certain amount, indicating a bias in the physics-based model. All three data science methods, NN, PGNN0, and PGNN, attempt to compensate for this bias by shifting their density profiles closer to the actual observations. On the three depth values where we have observations, we can see that both PGNN and PGNN0 show lower discrepancy with observations as compared to PHY. In fact, the density profile of PGNN matches almost perfectly with the observations, thus demonstrating the value in using physics-based loss function for better generalizability. However, the most striking insight from Figure \ref{fig:ml_profile} is that although the density estimate of PGNN0 is reasonably close to the three observations (thus indicating a low value of test RMSE), the density estimates soon start showing physically inconsistent patterns as we move lower in depth beyond the observations. In particular, the density estimates of PGNN0 start decreasing as we increase the depth beyond 6m. This is a violation of the monotonic relationship between density and depth as illustrated in Figure \ref{fig:density-depth}. The presence of such physical inconsistencies reduces the usefulness of a model's predictions in scientific analyses, even if the model shows low test RMSE. In contrast, the predictions of PGNN, while being closer to the actual observations, are always consistent with the monotonic relationship between density and depth.

Figure \ref{fig:men_profile} shows another example of density profiles on a different date in Lake Mendota. We can see that PGNN is again able to improve upon PHY and produce density estimates that are closest to the observations. On the other hand, both PGNN0 and NN shows large discrepancies with respect to the actual observations. This is because of the complex nature of relationships between the drivers and the temperature in Lake Mendota that are difficult to be captured without the use of physical relationships in the learning of neural networks. Additionally, the model predictions of PGNN0 can be seen to violate the physical relationship between density and depth (density estimates of PGNN0 decrease as we increase the depth from 10m to 12m), thus further reducing our confidence in PGNN0 representing physically meaningful results.

\section{Discussion on Alternate HPD Model Designs}
\label{sec:discussions}

\begin{figure}[tb]
\centering
\subfigure[Residual (Res) Model]{\label{fig:res} \includegraphics[width=0.7\textwidth]{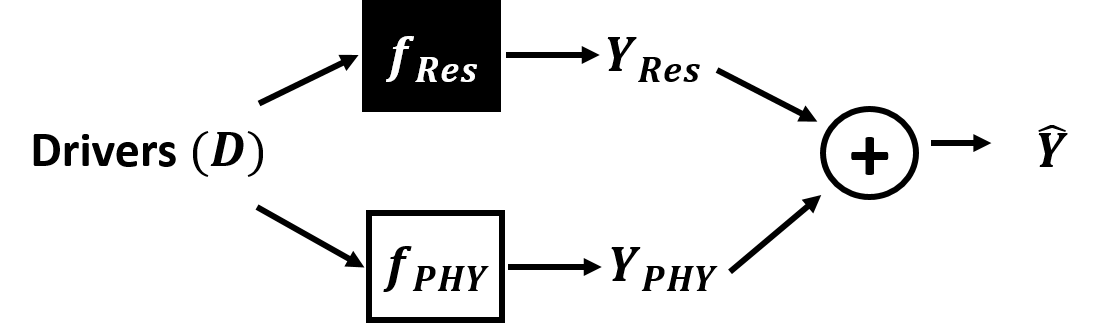}}
\\
\subfigure[Hybrid-Physics-Data-Residual (HPD-Res) Model]{\label{fig:hpdres} \includegraphics[width=0.7\textwidth]{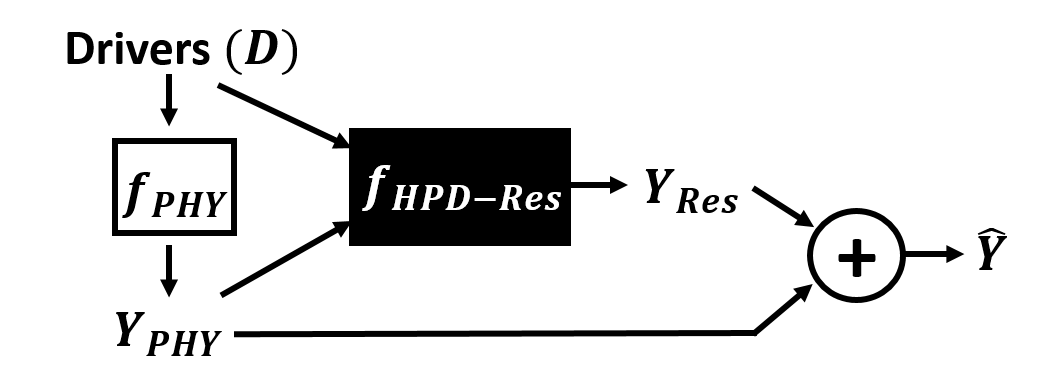}}
\caption{Alternate Hybrid-Physics-Data (HPD) model designs, where white boxes represent physics-based models while black boxes represent ML models.}
\label{fig:hybrid_arch}
\end{figure}

So far, we have demonstrated the value of hybrid-physics-data (HPD) modeling using a simple HPD design as illustrated in Figure \ref{fig:hpd}), where the outputs of the physics-based model are fed into the neural network model as additional features, along with the input drivers. In this section, we discuss its relevance in the context of two alternate HPD model designs based on residual modeling techniques (see Figure \ref{fig:hybrid_arch}), which are commonly used in the scientific literature to correct residuals of physics-based models using data-driven methods. The first HPD design (Figure \ref{fig:res}), termed the ``Residual Model,'' uses a simple ML model $f_{Res}$ to fix the residuals of physics-based model outputs $Y_{PHY}$ as additive correction terms. Specifically, instead of building an ML model to directly predict the target variable $Y$ from the input drivers $D$, we adopt a residual modeling strategy to predict $Y_{Res} (= Y-Y_{PHY})$, which when added to $Y_{PHY}$ provides corrected estimates of the target variable. Note that residual modeling is one of the simplest and most commonly used strategies for HPD modeling \cite{forssell1997combining,thompson1994modeling,san2018machine,san2018neural,wan2018data}. The primary motivation for building a residual model is to solve the simpler problem of estimating the residuals of a physics-based model, which are indicative of the systematic biases or equivalently the uncaptured variability of the physics-based model, instead of estimating the complete functional mapping from $D$ to $Y$. The final prediction of the target variable $Y$ is obtained by simply adding the predicted residual $Y_{Res}$ with the output of the physics model $Y_{PHY}$. In other words, a residual model can be thought of as a rectifying unit which aims to correct the predictions of the physics-based model. 

Another innovation in HPD design is illustrated in Figure \ref{fig:hpdres}, where the idea of residual modeling is combined with the idea of the basic HPD model described in Figure \ref{fig:hpd}. In this alternate HPD design, termed the ``Hybrid-Physics-Data-Residual (HPD-Res) Model,'' the ML model uses both the input drivers $D$ as well as the output of the physics-based models $Y_{PHY}$ to predict the residuals of the physics-based model $Y_{Res}$. The predicted residuals are then added to $Y_{PHY}$ to obtain the final predictions of the target variable $Y$. Note that HPD-Res shares some similarity with the basic residual (Res) model, as both of them predict the residual of the physics-based model instead of directly predicting the target variable. However, the difference in HPD-Res is that it uses $Y_{PHY}$ as additional inputs  in the ML architecture, which simplifies the task of learning the residuals (note that in some cases, it may be easier to identify patterns of systematic biases in the physics-based model by observing $D$ and $Y_{PHY}$ together). HPD-Res is also similar to the basic HPD model as both of them use $D$ and $Y_{PHY}$ as inputs in the ML model. However, the difference is that HPD-Res only predicts the residual $Y_{Res}$ to be added to $Y_{PHY}$ for deriving final predictions of the target variable $Y$. Hence, HPD-Res can be viewed as a `fusion' of the basic HPD and the basic Res models.

\begin{figure}[tb]
\centering
\subfigure[Lake Mendota]{\label{fig:mendotahybrid} \includegraphics[width=0.45\textwidth]{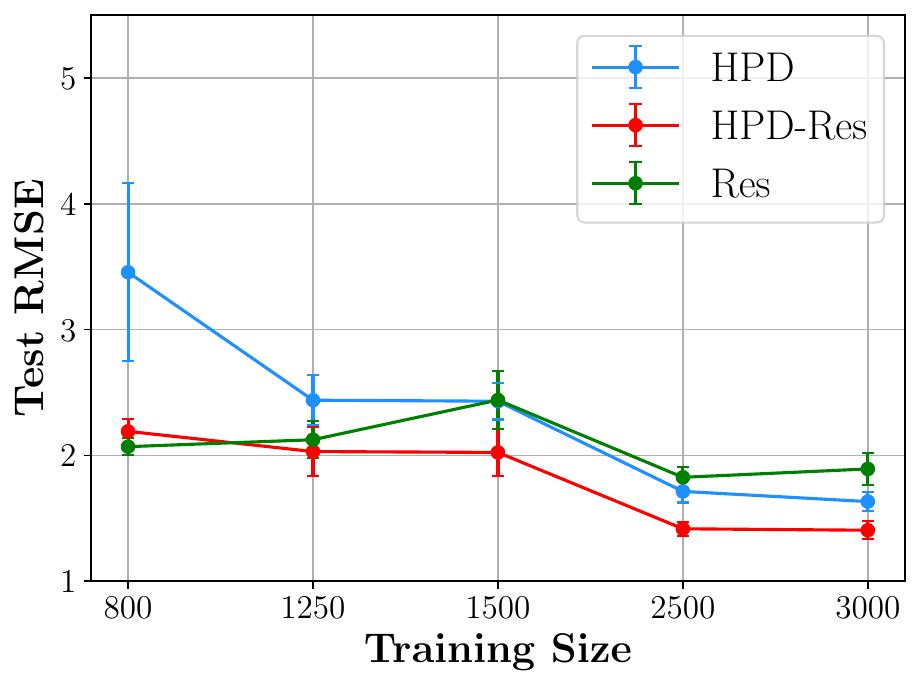}}
\subfigure[Mille Lacs Lake]{\label{fig:millelacshybrid} \includegraphics[width=0.45\textwidth]{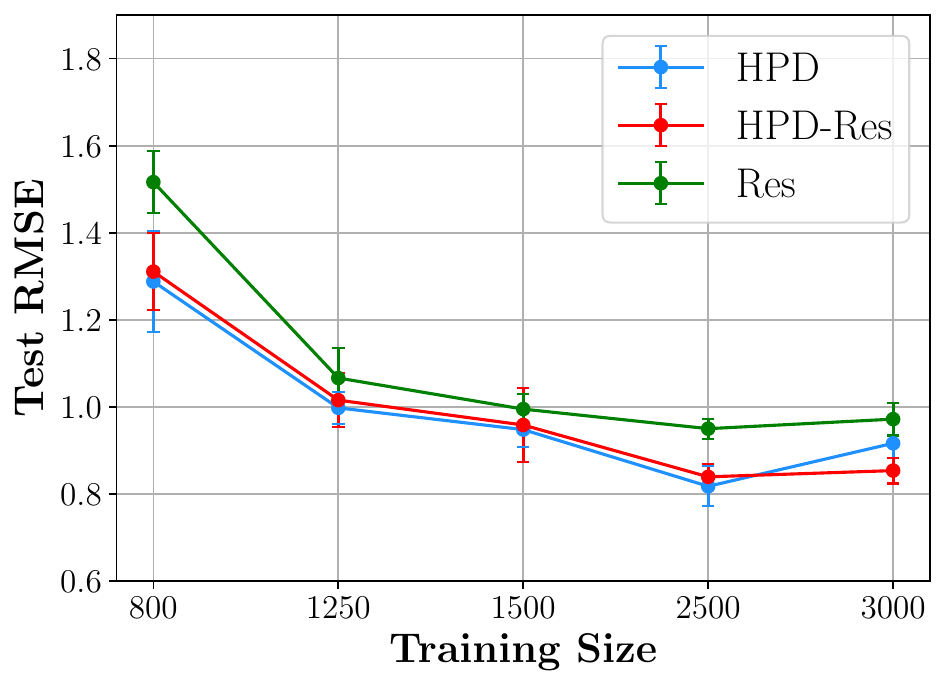}}
\caption{Comparing the performance of different hybrid-physics-data (HPD) model designs on Mille Lacs Lake and Lake Mendota at varying training sizes. Points and error bars respectively represent the mean and +/- one standard deviations from the mean of results from all 50 random weight initializations. The HPD model here corresponds to PGNN0 in Fig. \ref{fig:scatter}. Note that mean and standard deviations also vary from Fig. \ref{fig:scatter} due to different random weight initializations, and different versions of the Keras library used.} 
\label{fig:hybridmodels}
\end{figure}

To empirically understand the differences between the three HPD designs: basic HPD, basic Res, and HPD-Res, we compare their performances on Lake Mendota and Mille Lacs Lake at varying training sizes in Figure \ref{fig:hybridmodels}. Note that in these experiments, we did not include the physics-based loss function in the learning objective to solely evaluate the effect of HPD designs on generalization performance (as a result, the performance of the basic HPD model here corresponds to the PGNN0 baseline). We can see that across both lakes, the \emph{HPD-Res} performs slightly better than the basic HPD and the basic Residual formulations. In Lake Mendota, HPD-Res has a considerable difference in performance from HPD across all training sizes, and from Res at larger training sizes. On the other hand, in Mille Lacs Lake, the Res model performs the worst out of the three while HPD performs almost equivalently as HPD-Res. These results provide new insights on the differences between HPD model designs and suggests that further research on the choice of constructing HPD models is necessary. For example, one potential reason behind HPD-Res performing better than the basic HPD and the basic Res models is that HPD-Res combines the strengths of both these models; it uses the input drivers as well as $Y_{PHY}$ as inputs in the ML model, and the ML output is further added to $Y_{PHY}$ to correct its biases. Further research is needed to evaluate the validity of such claims regarding HPD model designs in different scientific problems involving a combination of physics knowledge and data. 


\section{Conclusions and Potential Future Work}
\label{sec:conclusions}

This paper presented a novel framework for learning physics-guided neural networks (PGNN), by using the outputs of physics-based model simulations as well as by leveraging physics-based loss functions to guide the learning of neural networks to physically consistent solutions. By anchoring neural network methods with scientific knowledge, we are able to show that the proposed framework not only shows better generalizability, but also produces physically meaningful results in comparison to black-box data science methods.

This paper serves as a stepping stone in the broader theme of research on using physics-based learning objectives in the training of data science models. While the specific formulation of PGNN explored in this paper was developed for the example problem of modeling lake temperature, similar developments could be explored in a number of other scientific and engineering disciplines where known forms of physical relationships can be exploited as physics-based loss functions. This paper paves the way towards learning neural networks by not only improving their ability to solve a given task, but also being cognizant of the physical relationships of the model outputs with other tasks, thus producing a more holistic view of the physical problem.

There are a number of directions of future research that can be explored as a continuation of this work. First, for the specific problem of lake temperature modeling, given the spatial and temporal nature of the problem domain, a natural extension would be to exploit the spatial and temporal dependencies in the test instances, e.g., by using recurrent neural network based architectures. Second, the analysis of the physically consistent model predictions produced by PGNN could be used to investigate the modeling deficiencies of the baseline physics-based model in detail. Third, while this paper presented a simple way of constructing hybrid-physics-data (HPD) models where $Y_{PHY}$ was ingested as an input in the data science model, more complex ways of constructing HPD models where the physics-based and data science components are tightly coupled need to be explored.
Fourth, theoretical analyses studying the impact of introducing physics-based loss functions on the sample complexity or convergence guarantees need to be investigated. Fifth, the research direction of PGNN can be complemented with other related efforts on producing interpretable data science results. In particular, the use of physics-based equations for interpreting the results of data science methods needs to be explored. Finally, while this paper explored the use of physical relationships between temperature, density, and depth of water in the learning of multi-layer perceptrons, other forms of physical relationships in different neural network models can be explored as future work. Of particular value would be to develop generative models that are trained to not only capture the structure in the unlabeled data, but are also guided by physics-based models to discover and emulate the known laws of physics. The paradigm of PGNN, if effectively utilized, could help in combining the strengths of physics-based and data science models, and opening a novel era of scientific discovery based on both physics and data.

\vspace{1ex}
\noindent \textbf{Disclaimer:} 
Any use of trade, firm, or product names is for descriptive purposes only and does not imply endorsement by the U.S. Government.

\bibliographystyle{unsrt}
\bibliography{main}

\begin{thebibliography}{25}
\providecommand{\natexlab}[1]{#1}
\providecommand{\url}[1]{\texttt{#1}}
\expandafter\ifx\csname urlstyle\endcsname\relax
  \providecommand{\doi}[1]{doi: #1}\else
  \providecommand{\doi}{doi: \begingroup \urlstyle{rm}\Url}\fi

\bibitem[Appenzeller(2017)]{Appenzeller16}
Tim Appenzeller.
\newblock The scientists{\textquoteright} apprentice.
\newblock \emph{Science}, 357\penalty0 (6346):\penalty0 16--17, 2017.

\bibitem[Graham-Rowe et~al.(2008)Graham-Rowe, Goldston, Doctorow, Waldrop,
  Lynch, Frankel, Reid, Nelson, Howe, Rhee, et~al.]{graham2008big}
D~Graham-Rowe, D~Goldston, C~Doctorow, M~Waldrop, C~Lynch, F~Frankel, R~Reid,
  S~Nelson, D~Howe, SY~Rhee, et~al.
\newblock Big data: science in the petabyte era.
\newblock \emph{Nature}, 455\penalty0 (7209):\penalty0 8--9, 2008.

\bibitem[Jonathan et~al.(2011)Jonathan, Gerald, et~al.]{jonathan2011special}
TO~Jonathan, AM~Gerald, et~al.
\newblock Special issue: dealing with data.
\newblock \emph{Science}, 331\penalty0 (6018):\penalty0 639--806, 2011.

\bibitem[Sejnowski et~al.(2014)Sejnowski, Churchland, and
  Movshon]{sejnowski2014putting}
Terrence~J Sejnowski, Patricia~S Churchland, and J~Anthony Movshon.
\newblock Putting big data to good use in neuroscience.
\newblock \emph{Nature neuroscience}, 17\penalty0 (11):\penalty0 1440--1441,
  2014.

\bibitem[Gupta and Nearing(2014)]{gupta2014debates}
Hoshin~V Gupta and Grey~S Nearing.
\newblock Debates—the future of hydrological sciences: A (common) path
  forward? using models and data to learn: A systems theoretic perspective on
  the future of hydrological science.
\newblock \emph{Water Resources Research}, 50\penalty0 (6):\penalty0
  5351--5359, 2014.

\bibitem[Lall(2014)]{lall2014debates}
Upmanu Lall.
\newblock Debates—the future of hydrological sciences: A (common) path
  forward? one water. one world. many climes. many souls.
\newblock \emph{Water Resources Research}, 50\penalty0 (6):\penalty0
  5335--5341, 2014.

\bibitem[McDonnell and Beven(2014)]{mcdonnell2014debates}
Jeffrey~J McDonnell and Keith Beven.
\newblock Debates—the future of hydrological sciences: A (common) path
  forward? a call to action aimed at understanding velocities, celerities and
  residence time distributions of the headwater hydrograph.
\newblock \emph{Water Resources Research}, 50\penalty0 (6):\penalty0
  5342--5350, 2014.

\bibitem[Magnuson et~al.(1979)Magnuson, Crowder, and
  Medvick]{magnuson1979temperature}
John~J Magnuson, Larry~B Crowder, and Patricia~A Medvick.
\newblock Temperature as an ecological resource.
\newblock \emph{American Zoologist}, 19\penalty0 (1):\penalty0 331--343, 1979.

\bibitem[Roberts et~al.(2013)Roberts, Fausch, Peterson, and
  Hooten]{roberts2013fragmentation}
James~J Roberts, Kurt~D Fausch, Douglas~P Peterson, and Mevin~B Hooten.
\newblock Fragmentation and thermal risks from climate change interact to
  affect persistence of native trout in the colorado river basin.
\newblock \emph{Global Change Biology}, 19\penalty0 (5):\penalty0 1383--1398,
  2013.

\bibitem[Rahel and Olden(2008)]{rahel2008assessing}
Frank~J Rahel and Julian~D Olden.
\newblock Assessing the effects of climate change on aquatic invasive species.
\newblock \emph{Conservation biology}, 22\penalty0 (3):\penalty0 521--533,
  2008.

\bibitem[Roberts et~al.(2017)Roberts, Fausch, Hooten, and
  Peterson]{roberts2017nonnative}
James~J Roberts, Kurt~D Fausch, Mevin~B Hooten, and Douglas~P Peterson.
\newblock Nonnative trout invasions combined with climate change threaten
  persistence of isolated cutthroat trout populations in the southern rocky
  mountains.
\newblock \emph{North American Journal of Fisheries Management}, 37\penalty0
  (2):\penalty0 314--325, 2017.

\bibitem[Harris and Graham(2017)]{harris2017predicting}
Ted~D Harris and Jennifer~L Graham.
\newblock Predicting cyanobacterial abundance, microcystin, and geosmin in a
  eutrophic drinking-water reservoir using a 14-year dataset.
\newblock \emph{Lake and Reservoir Management}, 33\penalty0 (1):\penalty0
  32--48, 2017.

\bibitem[Paerl and Huisman(2008)]{paerl2008blooms}
Hans~W Paerl and Jef Huisman.
\newblock Blooms like it hot.
\newblock \emph{Science}, 320\penalty0 (5872):\penalty0 57--58, 2008.

\bibitem[Hipsey et~al.(2014)Hipsey, Bruce, and Hamilton]{hipsey2014glm}
MR~Hipsey, LC~Bruce, and DP~Hamilton.
\newblock Glm—general lake model: Model overview and user information.
\newblock \emph{Perth (Australia): University of Western Australia Technical
  Manual}, 2014.

\bibitem[Martin and McCutcheon(1998)]{martin1998hydrodynamics}
James~L Martin and Steven~C McCutcheon.
\newblock \emph{Hydrodynamics and transport for water quality modeling}.
\newblock CRC Press, 1998.

\bibitem[Read et~al.(2017)Read, Carr, De~Cicco, Dugan, Hanson, Hart, Kreft,
  Read, and Winslow]{read2017water}
Emily~K Read, Lindsay Carr, Laura De~Cicco, Hilary~A Dugan, Paul~C Hanson,
  Julia~A Hart, James Kreft, Jordan~S Read, and Luke~A Winslow.
\newblock Water quality data for national-scale aquatic research: The water
  quality portal.
\newblock \emph{Water Resources Research}, 53\penalty0 (2):\penalty0
  1735--1745, 2017.

\bibitem[Prentice et~al.(1992)Prentice, Cramer, Harrison, Leemans, Monserud,
  and Solomon]{prentice1992special}
I~Colin Prentice, Wolfgang Cramer, Sandy~P Harrison, Rik Leemans, Robert~A
  Monserud, and Allen~M Solomon.
\newblock Special paper: a global biome model based on plant physiology and
  dominance, soil properties and climate.
\newblock \emph{Journal of biogeography}, pages 117--134, 1992.

\bibitem[Chollet(2015)]{chollet2015}
François Chollet.
\newblock keras.
\newblock \url{https://github.com/fchollet/keras}, 2015.

\bibitem[Zeiler(2012)]{zeiler2012adadelta}
Matthew~D Zeiler.
\newblock Adadelta: an adaptive learning rate method.
\newblock \emph{arXiv preprint arXiv:1212.5701}, 2012.

\bibitem[Fang et~al.(2012)Fang, Alam, Stefan, Jiang, Jacobson, and
  Pereira]{fang2012simulations}
Xing Fang, Shoeb~R Alam, Heinz~G Stefan, Liping Jiang, Peter~C Jacobson, and
  Donald~L Pereira.
\newblock Simulations of water quality and oxythermal cisco habitat in
  minnesota lakes under past and future climate scenarios.
\newblock \emph{Water Quality Research Journal}, 47\penalty0 (3-4):\penalty0
  375--388, 2012.

\bibitem[Forssell and Lindskog(1997)]{forssell1997combining}
Urban Forssell and Peter Lindskog.
\newblock Combining semi-physical and neural network modeling: An example ofits
  usefulness.
\newblock \emph{IFAC Proceedings Volumes}, 30\penalty0 (11):\penalty0 767--770,
  1997.

\bibitem[Thompson and Kramer(1994)]{thompson1994modeling}
Michael~L Thompson and Mark~A Kramer.
\newblock Modeling chemical processes using prior knowledge and neural
  networks.
\newblock \emph{AIChE Journal}, 40\penalty0 (8):\penalty0 1328--1340, 1994.

\bibitem[San and Maulik(2018{\natexlab{a}})]{san2018machine}
Omer San and Romit Maulik.
\newblock Machine learning closures for model order reduction of thermal
  fluids.
\newblock \emph{Applied Mathematical Modelling}, 60:\penalty0 681--710,
  2018{\natexlab{a}}.

\bibitem[San and Maulik(2018{\natexlab{b}})]{san2018neural}
Omer San and Romit Maulik.
\newblock Neural network closures for nonlinear model order reduction.
\newblock \emph{Advances in Computational Mathematics}, 44\penalty0
  (6):\penalty0 1717--1750, 2018{\natexlab{b}}.

\bibitem[Wan et~al.(2018)Wan, Vlachas, Koumoutsakos, and Sapsis]{wan2018data}
Zhong~Yi Wan, Pantelis Vlachas, Petros Koumoutsakos, and Themistoklis Sapsis.
\newblock Data-assisted reduced-order modeling of extreme events in complex
  dynamical systems.
\newblock \emph{PloS one}, 13\penalty0 (5):\penalty0 e0197704, 2018.

\end{thebibliography}

\end{document}